%% file: main.tex
\documentclass[10pt,twocolumn,letterpaper]{article}

\usepackage{cvpr}
\usepackage{times}
\usepackage{epsfig}
\usepackage{graphicx}
\usepackage{amsmath}
\usepackage{amssymb}
\usepackage{booktabs}
\usepackage{csvsimple}

\usepackage[pdfa,breaklinks=true,bookmarks=false,pagebackref=true,colorlinks,letterpaper=true]{hyperref}

\cvprfinalcopy 

\def\cvprPaperID{5828} 

\ifcvprfinal\fi
\begin{document}

\title{JSIS3D: Joint Semantic-Instance Segmentation of 3D Point Clouds with \\
  Multi-Task Pointwise Networks and Multi-Value Conditional Random Fields}

\author{
  Quang-Hieu Pham$^{\dagger}$ \hspace{0.2in}
  Duc Thanh Nguyen$^{\ddagger}$ \hspace{0.2in}
  Binh-Son Hua$^{\rhd}$ \hspace{0.2in}
  Gemma Roig$^{\dagger}$ \hspace{0.2in}
  Sai-Kit Yeung$^{\lhd}$
  \\
  $^{\dagger}$Singapore University of Technology and Design
  \\
  $^{\ddagger}$Deakin University \hspace{0.5in}
  $^{\rhd}$The University of Tokyo
  \\
  $^{\lhd}$Hong Kong University of Science and Technology
}

\maketitle
\thispagestyle{empty}

\begin{abstract}
  Deep learning techniques have become the to-go models for most vision-related
  tasks on 2D images. However, their power has not been fully realised on
  several tasks in 3D space, e.g., 3D scene understanding. In this work, we
  jointly address the problems of semantic and instance segmentation of 3D point
  clouds. Specifically, we develop a multi-task pointwise network that
  simultaneously performs two tasks: predicting the semantic classes of 3D
  points and embedding the points into high-dimensional vectors so that points
  of the same object instance are represented by similar embeddings. We then
  propose a multi-value conditional random field model to incorporate the
  semantic and instance labels and formulate the problem of semantic and
  instance segmentation as jointly optimising labels in the field model. The
  proposed method is thoroughly evaluated and compared with existing methods on
  different indoor scene datasets including S3DIS and SceneNN. Experimental
  results showed the robustness of the proposed joint semantic-instance
  segmentation scheme over its single components. Our method also achieved
  state-of-the-art performance on semantic segmentation.
\end{abstract}

\input{section/intro}
\input{section/related}
\input{section/method}
\input{section/eval}
\input{section/remark}

{\small
  \bibliographystyle{ieee}
  \bibliography{ref}
}

\end{document}

%% file: section/intro.tex
\section{Introduction}

\begin{figure*}[t]
    \centering
    \includegraphics[width=1.0\textwidth]{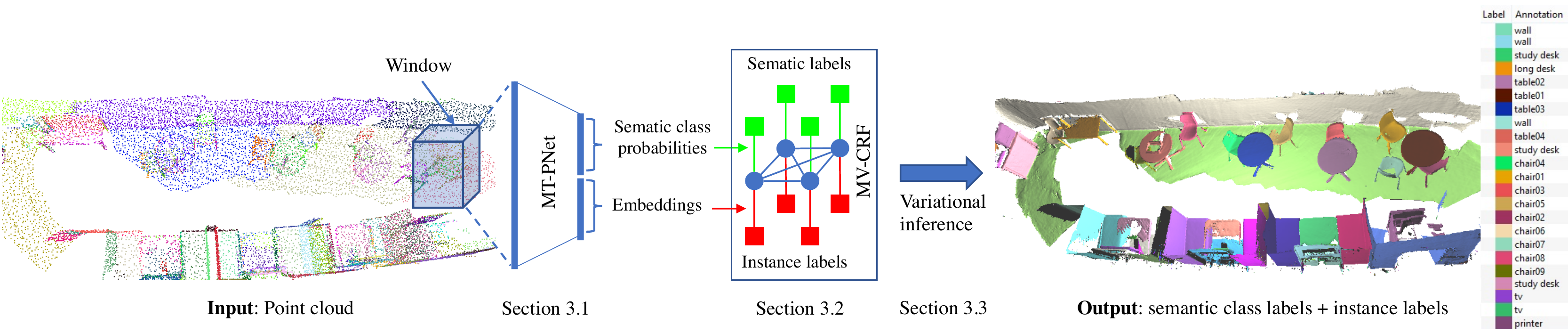}
    \caption{Pipeline of our proposed method. Given an input 3D point cloud, we
      scan the point cloud by overlapping windows. 3D vertices are then
      extracted from a window and passed through our multi-task neural network
      to get the semantic labels and instance embeddings. We then optimise a
      multi-value conditional random field model to produce the final
      results. Scene data is retrieved from \cite{hua-scenenn-3dv16}.}
    \label{fig:pipeline}
\end{figure*}

The growing popularity of low-cost 3D sensors (\eg, Kinect) and light-field
cameras has opened many 3D-based applications such as autonomous driving,
robotics, mobile-based navigation, virtual reality, and 3D games. This
development also acquires the capability of automatic understanding of 3D
data. In 2D domain, common scene understanding tasks including image
classification, semantic segmentation, or instance segmentation, have achieved
notable results \cite{he-maskrcnn-iccv17, chen-deeplab-pami18}. However, the
problem of 3D scene understanding poses much greater challenges, \eg,
large-scale and noisy data processing.

Literature has shown that the data of a 3D scene can be represented by a set of
images capturing the scene at different viewpoints \cite{hermans-dense-icra14,
  wolf-fast-icra15, vineet-incremental-icra15}, in a regular grid of volumes
\cite{wu-shapenets-cvpr15, maturana-voxnet-iros15, nguyen-field-cvpr16}, or
simply in a 3D point cloud \cite{qi-pointnet++-nips17, hua-pointwise-cvpr18,
  wang-dgcnn-arxiv18, huang-rsnet-cvpr18, li-pointcnn-nips18}. Our work is
inspired by the point-based representation for several reasons. Firstly,
compared with multi-view and volumetric representations, point clouds offer a
more compact and intuitive representation of 3D data. Secondly, recent neural
networks directly built on point clouds \cite{qi-pointnet++-nips17,
  hua-pointwise-cvpr18, li-pointcnn-nips18, wang-dgcnn-arxiv18,
  huang-rsnet-cvpr18, klokov-escape-iccv17, landrieu-superpoint-cvpr18,
  li-sonet-cvpr18, xu-spidercnn-eccv18} have shown promising results across
multiple tasks such as object recognition and semantic segmentation.

In this paper, we address two fundamental problems in 3D scene understanding:
semantic segmentation and instance segmentation. Semantic segmentation aims to
identify a class label or object category (\eg, chair, table) for every 3D point
in a scene while instance segmentation clusters the scene into object
instances. These two problems have often been tackled separately in which
instance segmentation/detection is a post-processing task of semantic
segmentation \cite{qi-frustum-cvpr18,pham-rpss-wacv19}. However, we have
observed that object categories and object instances are mutually dependent. For
instance, shape and appearance features extracted on an instance would help to
identify the object category of that instance. On the other hand, if two 3D
points are assigned to different object categories, they unlikely belong to the
same object instance. Therefore, it is desirable to couple semantic and instance
segmentation into a single task. Towards the above motivations, we make the
following contributions in our work.
\begin{itemize}
\item A network architecture namely multi-task pointwise network (MT-PNet) that
  simultaneously performs two tasks: predicting the object categories of 3D
  points in a point cloud, and embedding these 3D points into high-dimensional
  feature vectors that allow clustering the points into object instances.
\item A multi-value conditional random field (MV-CRF) model that formulates the
  joint optimisation of class labels and object instances into a unified
  framework, which can be efficiently solved using variational mean field
  technique. To the best of our knowledge, we are the first to explore the joint
  optimisation of semantics and instances in a unified framework.
\item Extensive experiments on different benchmark datasets to validate the
  proposed method as well as its main components. Experimental results showed
  that the joint semantic and instance segmentation outperformed each individual
  task, and the proposed method achieved state-of-the-art performance on
  semantic segmentation.
\end{itemize}

The remainder of the paper is organised as follows. Section \ref{sec:related}
briefly reviews related work. The proposed method is described in Section
\ref{sec:method}. Experiments and results are presented and discussed in Section
\ref{sec:experiments}. The paper is finally concluded in Section
\ref{sec:remark}.

%% file: section/related.tex
\section{Related Work}
\label{sec:related}

\begin{figure*}[t]
    \centering
    \includegraphics[width=1.0\textwidth]{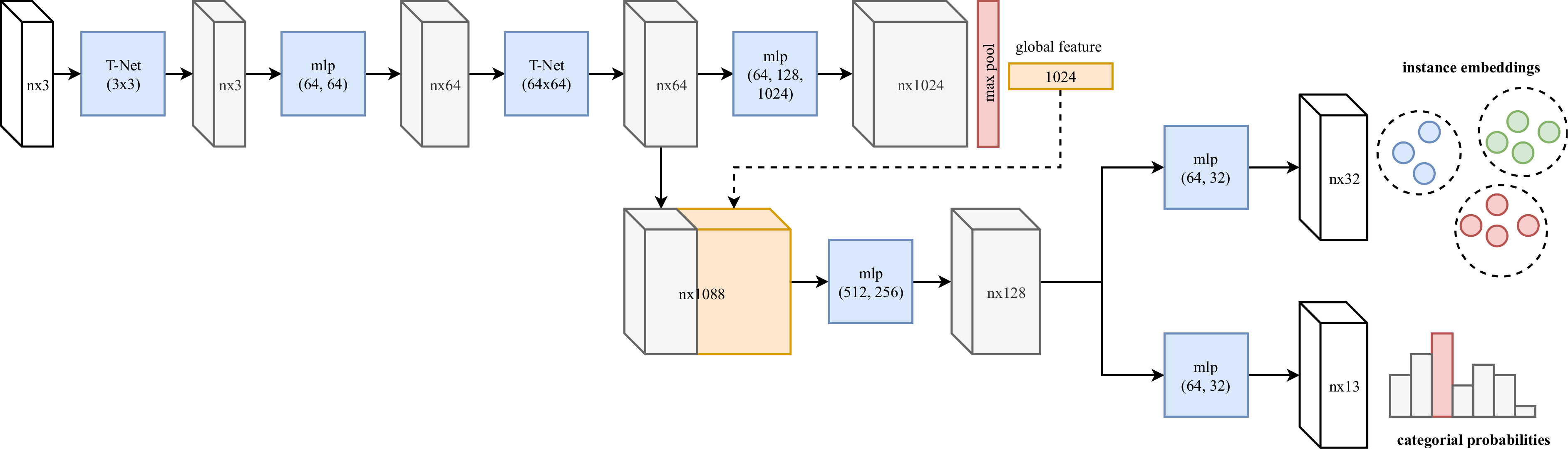}
    \caption{Our proposed MT-PNet architecture, which based on PointNet
      \cite{qi-pointnet-cvpr17}. The point cloud first go through a feed-forward
      neural network to compute a 128-dimension feature vector for each
      point. Here it splits into to branches: one for instance embedding and the
      other for semantic segmentation.}
    \label{fig:network}
\end{figure*}

This section reviews recent semantic and instance segmentation techniques in 3D
space. We especially focus on deep learning-based techniques applied on 3D point
clouds due to their proven robustness as well as being contemporary seminal in
the field. For the sake of brevity, we later refer to the traditional,
category-based semantic segmentation as \emph{semantic segmentation}, and
instance-based semantic segmentation as \emph{instance segmentation}.

\subsection{Semantic Segmentation}

Recent availability of indoor scene datasets \cite{silberman-nyud-eccv12,
  hua-scenenn-3dv16, dai-scannet-cvpr17, armeni-s3dis-cvpr16} has sparked
research interests in 3D scene understanding, particularly semantic
segmentation. We categorise these recent works into three main categories based
on their type of input data, namely multi-view images, volumetric
representation, and point clouds.

\paragraph{Multi-view approach.}
This approach often uses pre-trained models on 2D domain and applies them to 3D
space. Per-vertex labels are obtained by back-projecting and fusing 2D
predictions from colour or RGB-D images onto 3D space. Predictions on 2D can be
done via classifiers, \eg, random forests \cite{hermans-dense-icra14,
  riemenschneider-multiview-eccv14, wolf-fast-icra15,
  vineet-incremental-icra15}, or deep neural networks
\cite{mccormac-semanticfusion-icra17, yang-occupancy-iros17,
  pham-rpss-wacv19}. Such techniques can be implemented in tandem with 3D scene
reconstruction, creating a real-time semantic reconstruction system. However,
this approach suffers from inconsistencies between 2D predictions, and its
performance might depend on view placements.

\paragraph{Volumetric approach.}
The robustness of deep neural networks in solving several scene understanding
tasks on images has inspired applying deep neural networks directly in 3D space
to solve 3D scene understanding problem. In fact, convolutions on a regular
grid, \eg, image structures, can be easily extended to 3D, which leads to deep
learning with volumetric representation \cite{wu-shapenets-cvpr15,
  maturana-voxnet-iros15, nguyen-field-cvpr16}. To support high-resolution
segmentation and reduce memory footprints, a hierarchical data structure such as
an octree was proposed to limit convolution operations only on free-space voxels
\cite{riegler-octnet-cvpr17}. It has been shown that the performance of semantic
segmentation can be improved by solving the problem jointly with scene
completion \cite{song-sscnet-cvpr17, dai-scancomplete-cvpr18}.

\paragraph{Point cloud approach.}
In contrast to volume, point cloud is a compact yet intuitive representation
that directly stores attributes of the geometry of a 3D scene via coordinates
and normals of vertices. Point clouds arise naturally from commodity devices
such as multi-view stereos, depth, and LIDAR sensors. Point clouds can also be
converted to other representations such as volumes \cite{tchapmi-segcloud-3dv17}
or mesh \cite{valentin-mesh-cvpr13}. While convolutions can be done conveniently
on volumes \cite{tchapmi-segcloud-3dv17}, they are not applicable
straightforwardly on point clouds. This problem was first addressed in the work
of Qi \etal \cite{qi-pointnet-cvpr17}, and subsequently explored by several
others, \eg, \cite{qi-pointnet++-nips17, hua-pointwise-cvpr18,
  wang-dgcnn-arxiv18, huang-rsnet-cvpr18, li-pointcnn-nips18, li-sonet-cvpr18,
  xu-spidercnn-eccv18}. Semantic segmentation can further be extended to graph
convolution to handle large-scale point clouds
\cite{landrieu-superpoint-cvpr18}, and with the use of kd-tree to address
non-uniform point distributions \cite{klokov-escape-iccv17,
  groh-flexconv-accv18}.

\paragraph{Conditional Random Fields (CRFs)}
CRFs are often used in semantic segmentation of 3D scenes, \eg,
\cite{valentin-mesh-cvpr13, hermans-dense-icra14, kundu-monocular-eccv14,
  wolf-fast-icra15, vineet-incremental-icra15, mccormac-semanticfusion-icra17,
  qi-3dgraph-iccv17}. In general, CRFs make use of unary and binary potentials
capturing characteristics of individual 3D points \cite{wolf-fast-icra15} or
meshes \cite{valentin-mesh-cvpr13}, and their co-occurrence. To enhance CRFs
with prior knowledge, higher-order potentials are introduced
\cite{ladicky-what-eccv10, fidler-describing-cvpr12, zhu-cosegmentation-wacv14,
  arnab-hocrf-eccv16, yang-occupancy-iros17, feng-urban-eccv18,
  pham-rpss-wacv19}. Higher-order potentials, \eg, object detections
\cite{ladicky-what-eccv10, arnab-hocrf-eccv16, pham-rpss-wacv19}, act as
additional cues to help the inference of semantic class labels in CRFs.

\subsection{Instance Segmentation}
In general, there are two common strategies to tackle instance segmentation. The
first strategy is to localise object bounding boxes using object detection
techniques, and then find a mask that separates foreground and background within
each box. This approach has been shown to work robustly with images
\cite{dai-instance-cvpr16, he-maskrcnn-iccv17}, while deemed challenging in 3D
domain. This probably due to existing 3D object detectors are often not trained
from scratch but make use of image features \cite{deng-amodal-cvpr17,
  qi-frustum-cvpr18, liang-fusion-eccv18}. Extending such approaches with masks
is possible but might lead to a sub-optimal and more complicated pipeline.

Instead, given the promising results of semantic segmentation on 3D data
\cite{qi-pointnet-cvpr17, armeni-s3dis-cvpr16, hua-pointwise-cvpr18}, the
second strategy is to extend a semantic segmentation framework by adding a
procedure that proposes object instances. In an early attempt, Wang \etal
\cite{wang-sgpn-cvpr18} proposed to learn a semantic map and a similarity matrix
of point features based on the PointNet in \cite{qi-pointnet-cvpr17}. Authors
then proposed an heuristic and non-maximal suppression step to merge similar
points into instances.

%% file: section/method.tex
\section{Proposed Method}
\label{sec:method}

In this section, we describe our proposed method for semantic and instance
segmentation of 3D point clouds. Given a 3D point cloud, we first scan the
entire point cloud by overlapping 3D windows. Each window (with its associated
3D vertices) is passed to a neural network for predicting the semantic class
labels of the vertices within the window and embedding the vertices into
high-dimensional vectors. To enable such tasks, we develop a multi-task
pointwise network (MT-PNet) that aims to predict an object class for every 3D
point in the scene and at the same time to embed the 3D point with its class
label information into a vector. The network encourages 3D points belonging to
the same object instance be pulled to each other while pushing those of
different object instances as far away from each other as possible. Those class
labels and embeddings are then fused into a multi-value conditional random field
(MV-CRF) model. The semantic and instance segmentation are finally performed
jointly using variational inference. We illustrate the pipeline of our method in
Figure \ref{fig:pipeline} and describe its main components in the following
sub-sections.

\subsection{Multi-Task Pointwise Network (MT-PNet)}
Our MT-PNet is based on the feed forward architecture of PointNet proposed by Qi
\etal in \cite{qi-pointnet-cvpr17} (see Figure \ref{fig:network}). Specifically,
for an input point cloud of size $N$, a feature map of size $N \times D$, where
$D$ is the dimension of features for each point, is first computed. The MT-PNet
then diverges into two different branches performing two tasks: predicting the
semantic labels for 3D points and creating their pointwise instance
embeddings. The loss of our MT-PNet is the sum of the losses of its two
branches,
\begin{align}
  \label{eq:loss}
  \mathcal{L} = \mathcal{L}_{prediction} + \mathcal{L}_{embedding}
\end{align}

The prediction loss $\mathcal{L}_{prediction}$ is defined by the cross-entropy
as usual. Inspired by the work in \cite{de-discriminative-arxiv17}, we employ a
discriminative function to present the embedding loss $\mathcal{L}_{embedding}$.
In particular, suppose that there are $K$ instances and $N_k, k\in\{1,...,K\}$
is the number of elements in the $k$-th instance,
$\mathbf{e}_j \in \mathbb{R}^d$ is the embedding of point $v_j$, and
$\boldsymbol\mu_k$ is the mean of embeddings in the $k$-th instance. The
embedding loss can be defined as follows,
\begin{align}
  \label{eq:discriminative}
  \mathcal{L}_{embedding} = \alpha \cdot \mathcal{L}_{pull} + \beta \cdot \mathcal{L}_{push} + \gamma \cdot \mathcal{L}_{reg}
\end{align}
where
\begin{align}
  \label{eq:pull}
  \mathcal{L}_{pull} = \frac{1}{K} \sum_{k=1}^K \frac{1}{N_k} \sum_{j=1}^{N_k} \left [ \left \Vert \boldsymbol\mu_k - \mathbf{e}_j \right \Vert_2 - \delta_v \right ]^2_+
\end{align}
\begin{align}
  \label{eq:push}
  \mathcal{L}_{push} = \frac{1}{K(K-1)} \sum_{k=1}^K \sum_{m=1, m \neq k}^K \left [2\delta_d - \left \Vert \boldsymbol\mu_k - \boldsymbol\mu_m \right \Vert_2 \right ]^2_+
\end{align}
\begin{align}
  \label{eq:reg}
  \mathcal{L}_{reg} = \frac{1}{K} \sum_{k=1}^K \left \Vert \boldsymbol\mu_k \right \Vert_2
\end{align}
where $[x]_+=\max(0,x)$, $\delta_v$ and $\delta_d$ are respectively the margins
for the pull loss $\mathcal{L}_{pull}$ and push loss $\mathcal{L}_{push}$. We
set $\alpha = \beta = 1$ and $\gamma = 0.001$ in our implementation.

A simple intuition for this embedding loss is that the pull loss
$\mathcal{L}_{pull}$ attracts embeddings towards the centroids, \ie,
$\boldsymbol\mu_k$, while the push loss $\mathcal{L}_{push}$ keeps these
centroids away from each other. The regularisation loss $\mathcal{L}_{reg}$ acts
as a small force that draws all centroids towards the origin. As shown in
\cite{de-discriminative-arxiv17}, if we set the margin $\delta_d > 2\delta_v$,
then each embedding will be closer to its own centroid than other centroids.

\subsection{Multi-Value Conditional Random Fields (MV-CRF)}

Let $V=\{v_1,..,v_N\}$ be the point cloud of a 3D scene obtained after 3D
reconstruction. Each 3D vertex $v_j$ in the point cloud is represented by its 3D
location $\mathbf{l}_j=[x_j, y_j, z_j]$, normal
$\mathbf{n}_j = [n_{j,x}, n_{j,y}, n_{j,z}]$, and colour
$\mathbf{c}_j=[c_{j,R}, c_{j,G}, c_{j,B}]$. By using the proposed MT-PNet, we
also obtain an embedding $\mathbf{e}_j \in \mathbb{R}^d$ for each point
$v_j$. Let $L^S=\{l^S_1, ..., l^S_N\}$ be the set of semantic labels that need
to be assigned to the point cloud $V$, where $l^S_j$ represent the semantic
class, \eg, chair, table, etc., of $v_j$. Similarly, let
$L^I=\{l^I_1, ..., l^I_N\}$ be the set of instance labels of $V$, \ie, all
vertices of the same object instance will have the same instance label
$l^I_j$. The labels $l^S_j$ and $l^I_j$ are random variables taking values in
$S$ and $I$ which are the set of semantic labels and instance labels
respectively. Note that $S$ is predefined while $I$ is unknown and needs to be
determined through instance segmentation.

We now consider each vertex $v_j \in V$ as a node in a graph, two arbitrary
nodes $v_j$, $v_k$ are connected by an undirected edge, and each vertex $v_j$ is
associated with its semantic and instance labels represented by the random
variables $l^S_j$ and $l^I_j$. Our graph defined over $V$, $L^S$, and $L^I$ is
named multi-value conditional random fields (MV-CRF); this is because each node
$v_j$ is associated to two labels $(l^S_j, l^I_j)$ taking values in
$S \times I$. The joint semantic-instance segmentation of the point cloud $V$
thus can be formulated via minimising the following energy function,
\begin{align}
  \label{eq:energy}
  E(L^S,L^I|V) = &\sum_{j} \varphi(l^S_j) + \sum_{(j,k),j<k} \varphi(l^S_j, l^S_k) \notag \\
                 &+ \sum_{j} \psi(l^I_j) + \sum_{(j,k),j<k} \psi(l^I_j, l^I_k) \notag \\
                 &+ \sum_{s \in S} \sum_{i \in I} \phi(s,i)
\end{align}

We note that our MV-CRF substantially differs from existing higher-order CRFs,
\eg, \cite{ladicky-what-eccv10, fidler-describing-cvpr12, arnab-hocrf-eccv16,
  pham-rpss-wacv19}. Specifically, in existing higher-order CRFs, higher-orders,
\eg object detections, are used as prior knowledge that helps to improve
segmentation. In contrast, our MV-CRF treats instance labels and semantic labels
equally as unknown and optimises them simultaneously.

The energy function $E(L^S,L^I|V)$ in (\ref{eq:energy}) involves in a number of
potentials that incorporate physical constraints (\eg, surface smoothness,
geometric proximity) and semantic constraints (\eg, shape consistency between
object class and instances) in both semantic and instance
labeling. Specifically, the unary potential $\varphi(l^S_j)$ is defined over the
semantic labels $l^S_j$ and computed directly from the classification score of
MT-PNet as,
\begin{align}
  \label{eq:unary/semantic}
  \varphi(l^S_j = s) \propto -\log p(v_j|l^S_j = s)
\end{align}
where $s$ is a possible class label in $S$ and $p(v_j|l^S_j=s)$ is the
probability (\eg, softmax value) that our network classifies $v_j$ to the
semantic class $s$.

We have found that vertices of the same object class often share the same
distribution of classification scores, \ie, $p(v_j|l^S_j)$. We thus model the
pairwise potential $\varphi(l^S_j, l^S_k)$ via the classification scores of both
$v_j$ and $v_k$. Specifically, we define,
\begin{align}
  \label{eq:pairwise/semantic}
  \varphi(l^S_j, l^S_k) = \omega_{j,k} \exp \bigg\{ -\frac{[p(v_j|l^S_j)-p(v_k|l^S_k)]^2}{2\theta^2} \bigg\}
\end{align}
where $\omega_{j,k}$ is obtained from the Pott compatibility as,
\begin{align}
  \label{eq:pott}
  \omega_{j,k} =
  \begin{cases}
    -1, & \mbox{if } l^{S/I}_j = l^{S/I}_k\\
    1, & \mbox{otherwise}.
  \end{cases}
\end{align}

The unary potential $\psi(l^I_j)$ enforces embeddings belonged to the same
instance to get as close to their mean embeddings as possible. Intuitively,
embeddings of the same instance are expected to convert to their modes in the
embedding space. Meanwhile, embeddings of different instances are encouraged to
diverge from each other. Specifically, suppose that the instance label set
$I=\{i_1,...,i_{K}\}$ includes $K$ instances. Suppose that the current
configuration of $L^I$ assigns all the vertices in $V$ into these $K$
instances. For each instance label $i \in I$, we define,
\begin{align}
  \label{eq:unary/instance}
  \psi(l^I_j=i) &= -\frac{\exp\bigg[-\frac{1}{2}(\mathbf{e}_j - \boldsymbol\mu_i)^{\top} \boldsymbol\Sigma_i^{-1}(\mathbf{e}_j - \boldsymbol\mu_i) \bigg]}{\sqrt{(2\pi)^d|\boldsymbol\Sigma_i|}} \notag \\
                &-\log \bigg[ \sum_{k} 1(l^I_k=i) \bigg]
\end{align}
where $\boldsymbol\mu_i$ and $\boldsymbol\Sigma_i$ respectively denote the mean
and covariance matrix of embeddings assigned to the label $i$, and $1(\cdot)$ is
an indicator.

The term $\sum_{k} 1(l^I_k=i)$ in (\ref{eq:unary/instance}) represents the area
of instance $i$ and is used to favour large instances. We have found that this
term could help to remove tiny instances caused by noise in the point cloud.

The pairwise potential of instance labels $\psi(l^I_j, l^I_k)$ captures
geometric properties of surfaces in object instances and is defined as a mixture
of Gaussians of the locations, normals, and colour of vertices $v_j$ and
$v_k$. In particular,
\begin{align}
  \label{eq:pairwise/instance}
  &\psi(l^I_j, l^I_k) = \notag \\
  &\omega_{j,k} \exp \bigg( -\frac{\|\mathbf{l}_j-\mathbf{l}_k\|^2_2}{2\lambda_1^2} -\frac{\|\mathbf{n}_j-\mathbf{n}_k\|^2_2}{2\lambda_2^2} -\frac{\|\mathbf{c}_j-\mathbf{c}_k\|^2_2}{2\lambda_3^2} \bigg)
\end{align}
where $\omega_{j,k}$ is presented in (\ref{eq:pott}).

The term $\phi(s,i)$ in (\ref{eq:energy}) associates the semantic-based
potentials with instance-based potentials and encourages the consistency between
semantic and instance labels. For instance, if two vertices are assigned to the
same object instance, they should be assigned to the same object
class. Technically, if we compute a histogram $h_i$ of frequencies of semantic
labels $s$ for all vertices of object instance $i$, we can define $\phi(s,i)$
based on the mutual information between $s$ and $i$ as,
\begin{align}
  \label{eq:higher-order}
  \phi(s,i) = - h_i(s) \log h_i(s)
\end{align}
where $h_i(s)$ is the frequency that semantic label $s$ occurs in vertices whose
instance label is $i$.

As shown in (\ref{eq:higher-order}), given an instance label $i$, the sum of
$\phi(s,i)$ over all semantic labels $s \in S$ is the information entropy of the
labels $s$ w.r.t. the object instance $i$, \ie,
$\sum_{s \in S} \phi(s,i)=-\sum_{s \in S} h_i(s) \log h_i(s)$. A good labeling,
therefore, should minimise such entropy, leading to low variation of semantic
labels within the same object instance. Since the energy $E(L^S,L^I|V)$ in
(\ref{eq:energy}) sums over all semantic labels $s$ and instance labels $i$, it
would favour highly consistent labelings.

\subsection{Variational Inference}

The minimisation of $E(L^S,L^I|V)$ in (\ref{eq:energy}) is equivalent to the
maximisation of the posterior conditional $p(L^S,L^I|V)$ which is intractable to
be solved using a naive implementation. In this paper, we adopt mean field
variational approach to solve this optimisation problem
\cite{wainwright-graphical-ftml08}. In general, the idea of mean field
variational inference is to approximate the probability distribution
$p(L^S,L^I|V)$ by a variational distribution $Q(L^S, L^I)$ that can be fully
factorised over all random variables in $(L^S, L^I)$, \ie,
$Q(L^S,L^I)=\prod_j Q_j(l^S_j,l^I_j)$.

However, the factorisation of $Q(L^S, L^I)$ over all pairs in $(L^S, L^I)$
induces a computational complexity of $|S|\times|I|$ per vertex. In addition,
since our proposed MV-CRF model is fully connected, message passing steps used
in conventional implementation of mean field approximation require quadratic
complexity in the number of random variables (\ie, $2N$). Fortunately, since our
pairwise potentials, defined in (\ref{eq:pairwise/semantic}) and
(\ref{eq:pairwise/instance}), are expressed in Gaussians, message passing steps
can be performed efficiently via applying convolution operations with Gaussian
filters on downsampled versions of $Q$, followed by upsampling
\cite{krahenbuhl-densecrf-nips11}. Truncated Gaussians can be also be used to
approximate these Gaussian filters to further speed up the message passing
process \cite{paris-bilateral-eccv06}.

We first assume that $L^S$ and $L^I$ are independent in the joint variational
distribution $Q(L^S, L^I)$, and hence $Q(L^S, L^I)$ can be decomposed as,
\begin{align}
  \label{eq:factorisation}
  Q(L^S,L^I)=\bigg[ \prod_{j=1}^N Q^S_j(l^S_j) \bigg] \bigg[ \prod_{j=1}^N Q^I_j(l^I_j) \bigg]
\end{align}

The assumption in (\ref{eq:factorisation}) allows us to derive mean field update
equations for semantic and instance variational distributions $Q^S$ and $Q^L$.

Since the term $\sum_{s \in S} \sum_{i \in I} \phi(s,i)$ in (\ref{eq:energy}) is
not expressed in relative to the index $j$, for convenience to the computation
of mean field updates, for each vertex $v_j$, we define a new term $m_j$ as,
\begin{align}
  \label{eq:auxilary}
  m_j = \frac{\sum_{s \in S} h_{l^I_j}(s) \log h_{l^I_j}(s)}{\sum_{v_k \in V} 1(l^I_k=l^I_j)}
\end{align}

By using $m_j$, the term $\sum_{s \in S} \sum_{i \in I} \phi(s,i)$ in
(\ref{eq:energy}) can be rewritten as,
\begin{align}
  \sum_{s \in S} \sum_{i \in I} \phi(s,i) = \sum_{v_j \in V} m_j
\end{align}

We then obtain mean field updates,
\begin{align}
  \label{eq:update/semantic}
  Q^S_j(l^S_j=s) \leftarrow &\frac{1}{Z_j} \exp \bigg[ -\varphi(l^S_j=s) \notag \\
                            &-\sum_{s' \in S} \sum_{k \neq j} Q^S_k(l^S_k=s')\varphi(l^S_j,l^S_k)
                              -m_j \bigg],
\end{align}
and
\begin{align}
  \label{eq:update/instance}
  Q^I_j(l^I_j=i) \leftarrow &\frac{1}{Z_j} \exp \bigg[ -\psi(l^I_j=i) \notag \\
                            &-\sum_{i' \in I} \sum_{k \neq j} Q^I_k(l^I_k=i')\psi(l^I_j,l^I_k)
                              -m_j \bigg]
\end{align}
where $Z_j$ is the partition function that makes $Q(L^S,L^I)$ a probability mass
function during the optimisation.

%% file: section/eval.tex
\section{Experiments}
\label{sec:experiments}

\subsection{Experimental Setup}

Our MT-PNet was implemented in PyTorch. We trained our network using the SGD
optimiser. The learning rate was set to $0.01$ and decay rate was set to $0.5$
after every 50 epochs. The training took 10 hours on a single NVIDIA TITAN X
graphics card.

For the joint optimisation of semantic and instance labeling, we initialised the
semantic and instance labels for 3D vertices as follows. Semantic labels with
associated classification scores were obtained directly from MT-PNet.
Embeddings for all 3D vertices were also extracted. Initial instance labels were
then determined by applying the mean shift algorithm
\cite{comaniciu-meanshift-pami02} on the embeddings. The bandwidth of mean shift
was set to the margin of the push force $\delta_d$ in (\ref{eq:push}). We set
$\delta_d = 1.5$ and found this setting achieved the best performance. In
addition, when setting the bandwidth to lower values, our performance will drop
due to over-segmentation. We note that the number of clusters generated by the
mean shift algorithm may be much larger than the true number of instances since
we allow over-segmentation. After the joint optimisation step, we only maintain
instances that pertain at least one vertex.

Input of our MT-PNet is a point cloud of 4,096 points. To handle large-scale
scenes, an input point cloud was divided into overlapping windows, each of which
roughly contains 4,096 points. Each window was fed to our MT-PNet to extract
instance embeddings. The embeddings from all the windows were merged using the
\texttt{BlockMerging} procedure in SGPN \cite{wang-sgpn-cvpr18}. Joint
optimisation was then applied on the entire scene. Finally, we employ
non-maximal suppression to yield the final semantic-instance predictions.

\subsection{Datasets}
We conducted all experiments on two datasets: S3DIS \cite{armeni-s3dis-cvpr16}
and SceneNN \cite{hua-scenenn-3dv16}. S3DIS is a 3D scene dataset that includes
large-scale scans of indoor spaces at building level. On this dataset, we
performed experiments at the provided disjoint spaces, which were typically
parsed to about 10--80 object instances. The objects were annotated with 13
categories. We followed the original train/test split in
\cite{armeni-s3dis-cvpr16}. Since S3DIS does not include normals of 3D
vertices, we simplified (\ref{eq:pairwise/instance}) with only location and
colour.

SceneNN \cite{hua-scenenn-3dv16} is a scene meshes dataset of indoor scenes with
cluttered objects at room scale. Their semantic segmentation follows NYU-D v2
\cite{silberman-nyud-eccv12} category set, which has 40 semantic classes. On
this dataset, we followed the train/test split by Hua \etal
\cite{hua-pointwise-cvpr18}. Similar to S3DIS, the semantic and instance
segmentation were done on overlapping windows.

\subsection{Evaluation and Comparison}
\begin{figure*}[t!]
  \centering
  \begin{minipage}{\linewidth}
    S3DIS
  \end{minipage}\\
  \begin{minipage}{0.2\linewidth}
    \includegraphics[width=\linewidth]{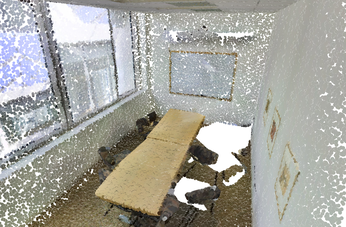}
  \end{minipage}%
  \begin{minipage}{0.2\linewidth}
    \includegraphics[width=\linewidth]{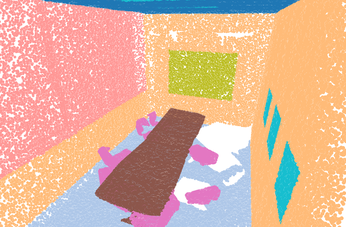}
  \end{minipage}%
  \begin{minipage}{0.2\linewidth}
    \includegraphics[width=\linewidth]{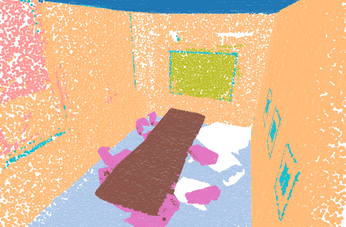}
  \end{minipage}%
  \begin{minipage}{0.2\linewidth}
    \includegraphics[width=\linewidth]{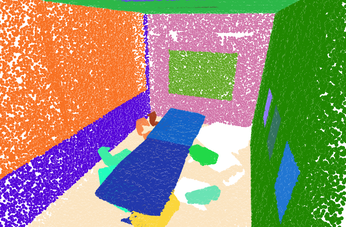}
  \end{minipage}%
  \begin{minipage}{0.2\linewidth}
    \includegraphics[width=\linewidth]{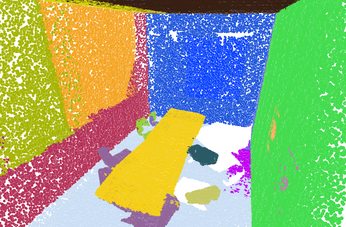}
  \end{minipage}\\
  \begin{minipage}{0.2\linewidth}
    \includegraphics[width=\linewidth]{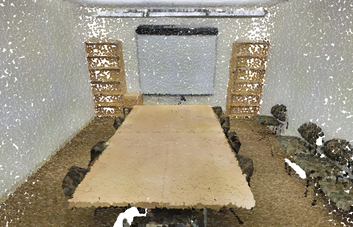}
  \end{minipage}%
  \begin{minipage}{0.2\linewidth}
    \includegraphics[width=\linewidth]{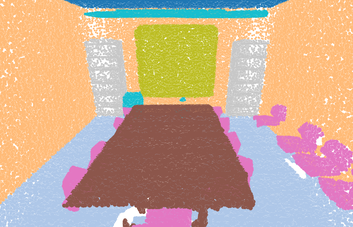}
  \end{minipage}%
  \begin{minipage}{0.2\linewidth}
    \includegraphics[width=\linewidth]{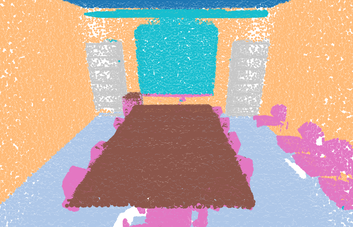}
  \end{minipage}%
  \begin{minipage}{0.2\linewidth}
    \includegraphics[width=\linewidth]{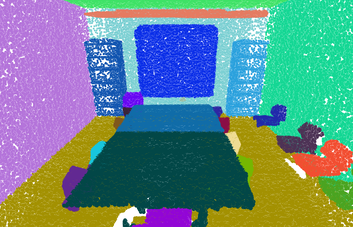}
  \end{minipage}%
  \begin{minipage}{0.2\linewidth}
    \includegraphics[width=\linewidth]{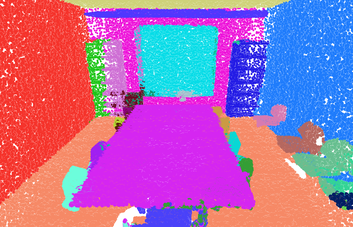}
  \end{minipage}\\
  \vspace{0.1in}
  \begin{minipage}{\linewidth}
    SceneNN
  \end{minipage}\\
  \begin{minipage}{0.2\linewidth}
    \includegraphics[width=\linewidth]{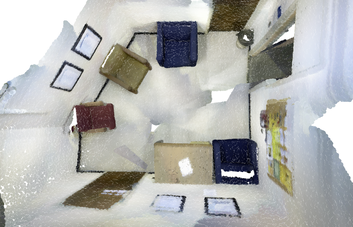}
  \end{minipage}%
  \begin{minipage}{0.2\linewidth}
    \includegraphics[width=\linewidth]{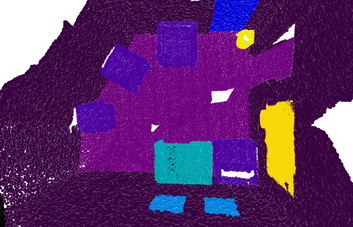}
  \end{minipage}%
  \begin{minipage}{0.2\linewidth}
    \includegraphics[width=\linewidth]{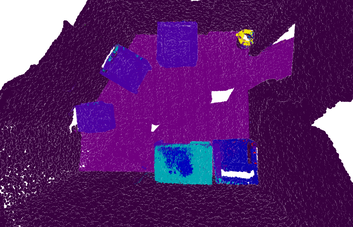}
  \end{minipage}%
  \begin{minipage}{0.2\linewidth}
    \includegraphics[width=\linewidth]{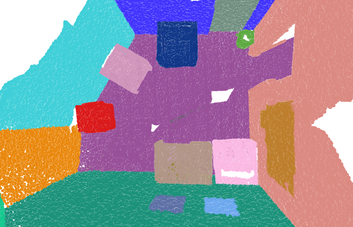}
  \end{minipage}%
  \begin{minipage}{0.2\linewidth}
    \includegraphics[width=\linewidth]{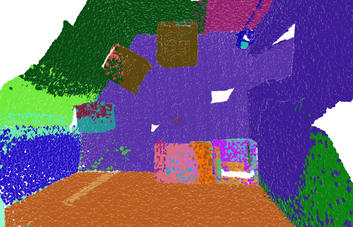}
  \end{minipage}\\
  \begin{minipage}{0.2\linewidth}
    \includegraphics[width=\linewidth]{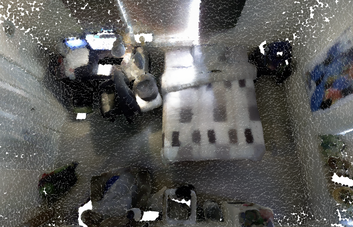}
  \end{minipage}%
  \begin{minipage}{0.2\linewidth}
    \includegraphics[width=\linewidth]{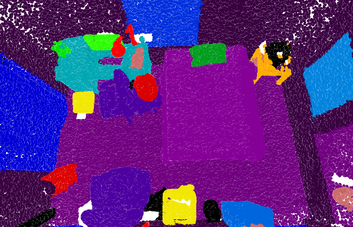}
  \end{minipage}%
  \begin{minipage}{0.2\linewidth}
    \includegraphics[width=\linewidth]{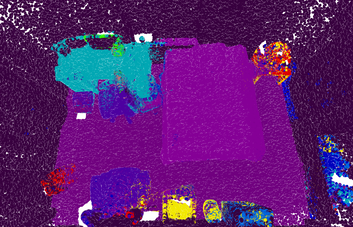}
  \end{minipage}%
  \begin{minipage}{0.2\linewidth}
    \includegraphics[width=\linewidth]{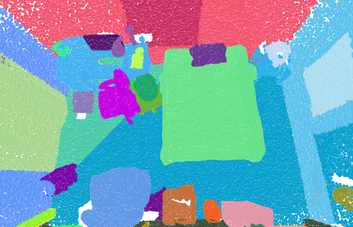}
  \end{minipage}%
  \begin{minipage}{0.2\linewidth}
    \includegraphics[width=\linewidth]{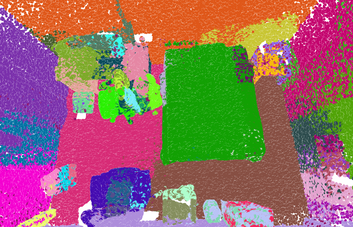}
  \end{minipage}
  \vspace{0.1in}
  \caption{Semantic and instance segmentation results. From left to right: input
    point cloud, ground truth of semantic segmentation, our semantic
    segmentation result, ground truth of instance segmentation, our instance
    segmentation result. For semantic segmentation, different colours represent
    different categories. For instance segmentation, different colours represent
    different instances.}
  \label{fig:qualitative}
\end{figure*}

In this section, we provide a comprehensive evaluation of our method and its
variants, and comparisons with existing methods in both semantic and instance
segmentation tasks. Several results of our method are shown in Figure
\ref{fig:qualitative}.

\paragraph{Ablation study.}

We study the effectiveness of joint semantic-instance segmentation compared with
its individual tasks. This study is done by investigating the role of potentials
of the energy of our MV-CRF defined in (\ref{eq:energy}). Specifically, for
semantic segmentation, we investigate the use of unary potentials in
(\ref{eq:unary/semantic}) only and traditional CRFs combining
(\ref{eq:unary/semantic}) and (\ref{eq:pairwise/semantic}). Similarly, for
instance segmentation, we compare the use of (\ref{eq:unary/instance}) only and
the combination of (\ref{eq:unary/instance}) and (\ref{eq:pairwise/instance}).
We also measure the performance of the joint task, \ie, the whole energy of
MV-CRF. Table \ref{tab:ablation} compares MV-CRF and its variants in both
semantic and instance segmentation on S3DIS. Metrics include micro-mean accuracy
(mAcc)\footnote{Micro-mean takes into account the size of classes in calculating
  the average accuracy and thus is often used for unbalanced data. In our
  context, micro-mean accuracy is equivalent to the overall accuracy that is
  often used in semantic segmentation.} \cite{sokolova-classification-ipm09} for
semantic segmentation and mAP@0.5 for instance segmentation.

\begin{table}[!ht]
  \begin{minipage}{.45\linewidth}
    \small
    \centering
    \begin{tabular}{rc}
      \multicolumn{2}{c}{Semantic segmentation} \\
      \toprule
      Method & mAcc \\
      \midrule
      (\ref{eq:unary/semantic}) & 86.7 \\
      (\ref{eq:unary/semantic}) + (\ref{eq:pairwise/semantic}) & 86.9 \\
      MV-CRF & \textbf{87.4} \\
      \bottomrule
    \end{tabular}
  \end{minipage}
  \begin{minipage}{.45\linewidth}
    \small
    \centering
    \begin{tabular}{rc}
      \multicolumn{2}{c}{Instance segmentation} \\
      \toprule
      Method & mAP@0.5 \\
      \midrule
      (\ref{eq:unary/instance}) & 24.9 \\
      (\ref{eq:unary/instance}) + (\ref{eq:pairwise/instance}) & 27.4 \\
      MV-CRF & \textbf{36.3} \\
      \bottomrule
    \end{tabular}
  \end{minipage}
  \caption{Comparison of our MV-CRF and its variants.}
  \label{tab:ablation}
\end{table}

\begin{table*}[t]
  \small
  \centering
  \begin{filecontents*}{figures/csv/s3dis/semantic.csv}
Method,mAcc,ceiling,floor,wall,beam,column,window,door,table,chair,sofa,bookcase,board,clutter
PointNet \cite{qi-pointnet-cvpr17},78.6,88.8,97.3,69.8,0.1,3.9,46.3,10.8,52.6,58.9,40.3,5.9,26.4,33.2
Pointwise \cite{hua-pointwise-cvpr18},81.5,97.9,99.3,92.7,\textbf{47.3},2.6,49.6,\textbf{50.6},74.1,58.2,0,39.3,0,61.1
SEGCloud \cite{tchapmi-segcloud-3dv17},80.8,90.1,96.1,69.9,0,\textbf{18.4},38.4,23.1,75.9,70.4,\textbf{58.4},40.9,13,41.6
Ours (MT-PNet),86.7,97.4,99.6,92.7,11.9,4.6,\textbf{60.1},26.4,\textbf{80.8},83.7,23.7,61.1,\textbf{55.2},\textbf{70.6}
Ours (MV-CRF),\textbf{87.4},\textbf{98.4},\textbf{99.6},\textbf{94.4},4.7,1.5,59.7,24.9,80.6,\textbf{84.9},30,\textbf{63.0},52.5,70.5
\end{filecontents*}
\csvreader[
  head to column names,
  tabular=rcccccccccccc,
  table head=
  \toprule
  Method & mAcc & ceiling & floor & wall &  window & door & table & chair & sofa & bookcase & board & clutter \\
  \midrule,
  table foot=\bottomrule
  ]{figures/csv/s3dis/semantic.csv}{
    Method=\Method,
    mAcc=\mAcc,
    ceiling=\ceiling,
    floor=\floor,
    wall=\wall,
    beam=\beam,
    column=\column,
    window=\window,
    door=\door,
    table=\table,
    chair=\chair,
    sofa=\sofa,
    bookcase=\bookcase,
    board=\board,
    clutter=\clutter
  }{
    \Method &
    \mAcc &
    \ceiling &
    \floor &
    \wall &
    \window &
    \door &
    \table &
    \chair &
    \sofa &
    \bookcase &
    \board &
    \clutter
  }
  \caption{Semantic segmentation results on S3DIS. Here we also show the
    stand-alone performance of MT-PNet, and when running the full pipeline with
    MV-CRF.}
  \label{tab:s3dis/semantic}
\end{table*}

\begin{table*}[t]
  \small
  \centering
  \begin{filecontents*}{figures/csv/s3dis/instance.csv}
Method,mAP,ceiling,floor,wall,beam,column,window,door,table,chair,sofa,bookcase,board,clutter
Armeni \etal \cite{armeni-s3dis-cvpr16},-,71.6,88.7,72.9,66.7,91.8,25.9,54.1,46,16.2,6.8,54.7,3.9,-
SGPN \cite{wang-sgpn-cvpr18},54.4,\textbf{79.4},66.3,\textbf{88.8},\textbf{78.0},\textbf{60.7},\textbf{66.6},\textbf{56.8},\textbf{46.9},\textbf{40.8},6.4,\textbf{47.6},11.1,-
Ours (MT-PNet),24.9,71.5,78.4,28.3,0,0,24.4,3.5,12.1,36.2,10,12.6,34.5,12.8
Ours (MV-CRF),36.3,76.9,\textbf{83.6},32.2,0,0,51.4,7.2,16.3,23.6,\textbf{16.7},21.8,\textbf{52.1},\textbf{13.4}
\end{filecontents*}
\csvreader[
  head to column names,
  tabular=rcccccccccccc,
  table head=
  \toprule
  Method & mAP & ceiling & floor & wall & window & door & table & chair & sofa & bookcase & board & clutter \\
  \midrule,
  table foot=\bottomrule
  ]{figures/csv/s3dis/instance.csv}{
    Method=\Method,
    mAP=\mAP,
    ceiling=\ceiling,
    floor=\floor,
    wall=\wall,
    window=\window,
    door=\door,
    table=\table,
    chair=\chair,
    sofa=\sofa,
    bookcase=\bookcase,
    board=\board,
    clutter=\clutter
  }{
    \Method &
    \mAP &
    \ceiling &
    \floor &
    \wall &
    \window &
    \door &
    \table &
    \chair &
    \sofa &
    \bookcase &
    \board &
    \clutter
  }
  \caption{Instance segmentation results on S3DIS. Here we also show the
    stand-alone performance of MT-PNet, and when running the full pipeline with
    MV-CRF. Note that results from Armeni \etal are on 3D bounding boxes instead
    of point clouds.}
  \label{tab:s3dis/instance}
\end{table*}

\begin{table*}[t]
  \small
  \centering
  \begin{filecontents*}{figures/csv/scenenn/semantic.csv}
Method,wall,floor,cabinet,bed,chair,sofa,table,door,window,desk,tv,prop
Pointwise \cite{hua-pointwise-cvpr18},93.8,88.6,1.5,11.6,58.6,5.5,23.5,0.0,0.0,29.5,7.7,5.8
Ours (MT-PNet),94.2,91.5,9.2,58.4,81.4,10.9,37.3,\textbf{4.1},0.0,54.0,\textbf{33.3},\textbf{13.2}
Ours (MLS-CRF),\textbf{96.0},\textbf{92.4},\textbf{10.0},\textbf{74.6},\textbf{83.0},\textbf{11.0},\textbf{44.5},0.3,0.0,\textbf{61.7},24.4,11.1
\end{filecontents*}
\csvreader[
  head to column names,
  tabular=rcccccccccc,
  table head=
  \toprule
  Method & wall & floor & cabinet & bed & chair & sofa & table & desk & tv & prop \\
  \midrule,
  table foot=\bottomrule
  ]{figures/csv/scenenn/semantic.csv}{
    Method=\Method,
    wall=\wall,
    floor=\floor,
    cabinet=\cabinet,
    bed=\bed,
    chair=\chair,
    sofa=\sofa,
    table=\table,
    door=\door,
    window=\window,
    desk=\desk,
    tv=\tv,
    prop=\prop
  }{
    \Method &
    \wall &
    \floor &
    \cabinet &
    \bed &
    \chair &
    \sofa &
    \table &
    \desk &
    \tv &
    \prop
  }
  \caption{Semantic segmentation results on SceneNN. Here we only show a subset
    of representative classes of NYUv2, as some of the classes are not presented
    in SceneNN.}
  \label{tab:scenenn/semantic}
\end{table*}

\begin{table*}[t]
  \small
  \centering
  \begin{filecontents*}{figures/csv/scenenn/instance.csv}
Method,wall,floor,cabinet,bed,chair,sofa,table,door,window,desk,tv,prop
Ours (MT-PNet),13.1,27.3,0.0,15.0,\textbf{21.2},0.0,0.7,0.0,0.0,0.0,6.0,\textbf{2.0}
Ours (MLS-CRF),\textbf{13.9},\textbf{44.5},0.0,\textbf{32.9},12.9,0.0,\textbf{5.7},0.0,0.0,\textbf{10.8},0.0,0.8
\end{filecontents*}
\csvreader[
  head to column names,
  tabular=rcccccccccccc,
  table head=
  \toprule
  Method & wall & floor & cabinet & bed & chair & sofa & table & desk & tv & prop\\
  \midrule,
  table foot=\bottomrule
  ]{figures/csv/scenenn/instance.csv}{
    Method=\Method,
    wall=\wall,
    floor=\floor,
    cabinet=\cabinet,
    bed=\bed,
    chair=\chair,
    sofa=\sofa,
    table=\table,
    door=\door,
    window=\window,
    desk=\desk,
    tv=\tv,
    prop=\prop
  }{
    \Method &
    \wall &
    \floor &
    \cabinet &
    \bed &
    \chair &
    \sofa &
    \table &
    \desk &
    \tv &
    \prop
  }
  \caption{Instance segmentation results on SceneNN. Here we only show a subset
    of representative classes of NYUv2, as some of the classes are not presented
    in SceneNN.}
  \label{tab:scenenn/instance}
\end{table*}

\begin{figure}[h!t]
  \centering
  \includegraphics[width=\linewidth]{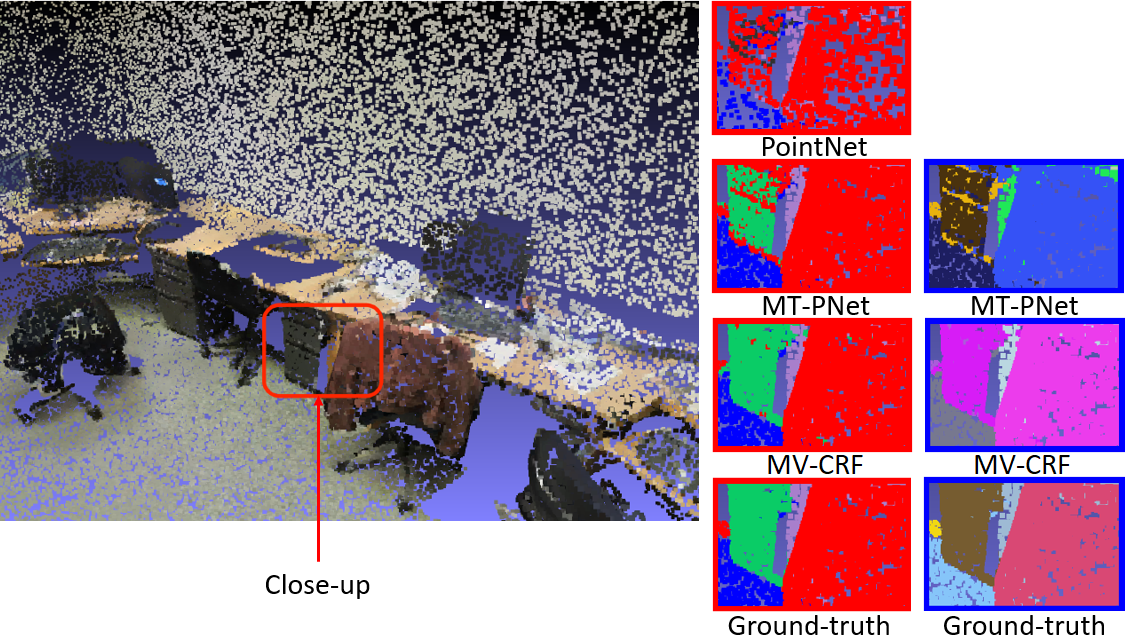}
  \caption{A close-up example of our method. Left: input, middle: semantic
    segmentation, right: instance segmentation.}
  \label{fig:closeup}
\end{figure}

\paragraph{Semantic segmentation.}
Table \ref{tab:s3dis/semantic} and Table \ref{tab:scenenn/semantic} show the
performance of our proposed method in semantic segmentation on S3DIS and SceneNN
dataset, respectively.

In this task, we evaluate the stand-alone performance of MT-PNet, marked as
``Ours (MT-PNet)'', and when running the full pipeline with MV-CRF, marked as
``Ours (MV-CRF)''. We also compare our method with other state-of-the-art deep
neural networks including PointNet \cite{qi-pointnet-cvpr17}, PointwiseCNN
\cite{hua-pointwise-cvpr18}, and SEGCloud \cite{tchapmi-segcloud-3dv17}. The
evaluation metric is per-class accuracy and micro-mean accuracy.

Experimental results show that our proposed MT-PNet significantly outperforms
its original architecture (\ie, PointNet \cite{qi-pointnet-cvpr17}), and the
improvement comes from the multi-task architecture. To confirm this, we
performed an experiment where we trained our MT-PNet with the instance embedding
branch disabled. The disabled-embedding branch network obtained the same
performance with the vanilla PointNet on semantic segmentation task.

As shown in Table \ref{tab:s3dis/semantic} and Table \ref{tab:scenenn/semantic},
our MV-CRF also well improves the base results from MT-PNet and achieves
state-of-the-art performance on semantic segmentation. This proves that
multi-task learning and joint optimisation can be beneficial. Figure
\ref{fig:closeup} shows a close-up example to illustrate the potential of our
MV-CRF in semantic segmentation.

\paragraph{Instance segmentation.}

We consider instance segmentation as object detection and thus evaluate this
task using average precision (AP) with IoU threshold at 0.5. To generate object
hypotheses, each instance $j$ is granted a confidence score $f_j$ calculated as,
\begin{align}
  \label{eq:confidence}
  f_j = &\frac{1}{\left | V_j \right |} \log \bigg\{ \prod_{v_k \in V_j} \bigg[ Q_k^S(l_k^S=s_j)Q_k^I(l_k^I=j) \bigg] \bigg\}
\end{align}
where $V_j$ is the set of points that have instance label $j$, and $Q_j^S$ and
$Q_j^L$ are defined in (\ref{eq:update/semantic}) and (\ref{eq:update/instance})
respectively.

Table \ref{tab:s3dis/instance} and Table \ref{tab:scenenn/instance} report the
instance segmentation performance of our method on S3DIS and SceneNN dataset,
respectively. We refer to the results obtained by applying the mean shift
algorithm directly on embeddings from MT-PNet as ``Ours (MT-PNet)'' and the
results of the full pipeline with MV-CRF as ``Ours (MV-CRF)''. Similarly to
semantic segmentation, experimental results show that our MV-CRF significantly
boosts up the segmentation performance in comparison to MT-PNet. Figure
\ref{fig:closeup} shows a qualitative comparison of our MV-CRF and other methods
in instance segmentation.

We also compare our method with other existing methods including SGPN
\cite{wang-sgpn-cvpr18}, a recent method for instance segmentation of point
clouds, and additional results from Armeni \etal \cite{armeni-s3dis-cvpr16}.
Compared with the state-of-the-art, our method shows clear improvement on some
categories, \eg, floor, sofa, board, and clutter. However, it produces low
precision segmentation results on other categories such as door. We have found
that this is mainly due to the low semantic segmentation accuracy in these
categories.

%% file: section/remark.tex
\section{Conclusion}
\label{sec:remark}

Semantic and instance segmentation of point clouds are crucial and fundamental
steps in 3D scene understanding. This paper proposes a semantic-instance
segmentation method that jointly performs both of the tasks via a novel
multi-task pointwise network and a multi-value conditional random field
model. The multi-task pointwise network simultaneously learns both the class
labels of 3D points and their embedded representations which enable clustering
3D points into object instances. The multi-value conditional random field model
integrates both 3D and high-dimensional embedded features to jointly perform
both semantic and instance segmentation. We evaluated the proposed method and
compared it with existing methods on different challenging indoor
datasets. Experimental results favourably showed the advance of our method in
comparison to state-of-the-art, and the joint semantic-instance segmentation
approach outperformed its individual components.

\paragraph{Acknowledgement.} This research project is partially supported by an
internal grant from HKUST (R9429) and the MOE SUTD SRG grant (SRG ISTD 2017
131).